\pdfoutput=1

\documentclass[11pt]{article}

\usepackage[]{acl}

\usepackage{times}
\usepackage{latexsym}

\usepackage[T1]{fontenc}

\usepackage[utf8]{inputenc}

\usepackage{microtype}

\usepackage[edges]{forest}
\usepackage[framemethod=tikz]{mdframed}
\usepackage{subcaption}
\usepackage{soul}
\definecolor{green}{rgb}{0.847,0.922,0.831}
\definecolor{lred}{rgb}{0.925, 0.816, 0.863}

\tikzset{%
    parent/.style =          {align=center,text width=2cm,rounded corners=3pt, line width=0.3mm, fill=gray!10,draw=gray!80},
    child/.style =           {align=center,text width=2.3cm,rounded corners=3pt, fill=blue!10,draw=blue!80,line width=0.3mm},
    grandchild/.style =      {align=center,text width=2cm,rounded corners=3pt},
    greatgrandchild/.style = {align=center,text width=1.5cm,rounded corners=3pt},
    greatgrandchild2/.style = {align=center,text width=1.5cm,rounded corners=3pt},    
    referenceblock/.style =  {align=center,text width=1.5cm,rounded corners=2pt},
    pretrain/.style =           {align=center,text width=1.8cm,rounded corners=3pt, fill=blue!10,draw=blue!80,line width=0.3mm},   
    pretrain_work/.style =           {align=center, text width=5cm,rounded corners=3pt, fill=blue!10,draw=blue!0,line width=0.3mm},  
    template/.style =           {align=center,text width=1.8cm,rounded corners=3pt, fill=red!10,draw=red!80,line width=0.3mm},   
    template_work/.style =           {align=center,text width=5cm,rounded corners=3pt, fill=red!10,draw=red!0,line width=0.3mm},    
    answer/.style =           {align=center,text width=1.8cm,rounded corners=3pt, fill= cyan!10,draw= cyan!80,line width=0.3mm},   
    answer_work/.style =           {align=center,text width=5cm,rounded corners=3pt, fill= cyan!10,draw= cyan!0,line width=0.3mm},      
    multiple/.style =           {align=center,text width=1.8cm,rounded corners=3pt, fill= orange!10,draw= orange!80,line width=0.3mm},   
    multiple_work/.style =           {align=center,text width=5cm,rounded corners=3pt, fill= orange!10,draw= orange!0,line width=0.3mm},        
    tuning/.style =           {align=center,text width=1.8cm,rounded corners=3pt, fill= magenta!10,draw= magenta!80,line width=0.3mm},   
    tuning_work/.style =           {align=center,text width=5cm,rounded corners=3pt, fill= magenta!10,draw= magenta!0,line width=0.3mm},          
}

\usepackage{tabularx}
\usepackage{xcolor}
\usepackage{colortbl}

\usepackage{booktabs}
\usepackage{amsmath}

\usepackage{geometry}
\usepackage{changepage}
\usepackage{amsmath, amsthm, amsfonts, amssymb}
\usepackage{thmtools}
\usepackage{mathtools}
\usepackage{xspace}
\usepackage{tcolorbox}
\usepackage{bm}

\usepackage{arydshln}
\usepackage{listings}
\newcommand{\lcell}[1]{\begin{tabular}{@{}l@{}}#1\end{tabular}}

\definecolor{gray}{rgb}{0.960,0.960,0.960}

\newcommand{\gsrow}[1]{\rowcolor{gray} {#1}}
\newcommand{\rg}[0]{\rowcolor{gray}}
\newcommand{\CC}[1]{\cellcolor[gray]{0.95}{#1}}
\newcommand{\WCC}[1]{\cellcolor[gray]{1.0}{#1}}

\usepackage{multirow}
\usepackage{hhline}
\usepackage{colortbl}
\usepackage{multicol}
\newcolumntype{C}[1]{>{\centering\arraybackslash}m{#1}}
\newcolumntype{L}[1]{>{\raggedright\arraybackslash}m{#1}}

\title{Recent Advances in Text-to-SQL:\\A Survey of What We Have and What We Expect}

\author{
    Naihao Deng \\
    University of Michigan\\
    \href{mailto:dnaihao@umich.edu}{dnaihao@umich.edu} \\
    \And 
    Yulong Chen \\
    Westlake University\\
    \href{mailto:yulongchen1010@gmail.com}{yulongchen1010@gmail.com} \\
    \And 
    Yue Zhang \\
    Westlake University \\
    \href{mailto:yue.zhang@wias.org.cn}{yue.zhang@wias.org.cn} \\
}

\begin{document}
\maketitle
\begin{abstract}

Text-to-SQL has attracted attention from both the natural language processing and database communities because of its ability to convert the semantics in natural language into SQL queries and its practical application in building natural language interfaces to database systems.
The major challenges in text-to-SQL lie in encoding the meaning of natural utterances, decoding to SQL queries, and translating the semantics between these two forms.
These challenges have been addressed to different extents by the recent advances.
However, there is still a lack of comprehensive surveys for this task.
To this end, we review recent progress on text-to-SQL for datasets, methods, and evaluation and provide this systematic survey, addressing the aforementioned challenges and discussing potential future directions.
We hope that this survey can serve as quick access to existing work and motivate future research. \footnote{The Github Link for this survey is: \url{https://github.com/text-to-sql-survey-coling22/text-to-sql-survey-coling22.github.io}.}
\end{abstract}


\section{Introduction}

\begin{figure}[t!]
    \centering
    \includegraphics[width=0.45\textwidth]{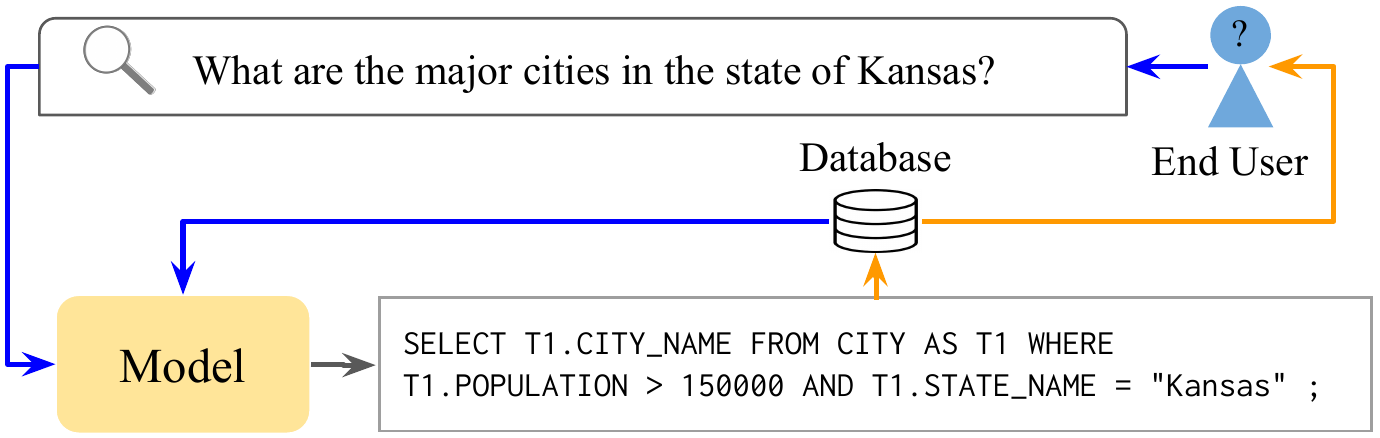}
    \caption{The framework for text-to-SQL systems. Given the database schema and user utterance, the system outputs a corresponding SQL query to query the database system for the result. Appendix~\ref{sec: text-to-sql-examples} gives more text-to-SQL examples.}
    \label{tab: text-to-SQL-framework}
\end{figure}

The task of text-to-SQL is to convert natural utterances into SQL queries~\cite{zhong2017seq2sql, yu2018spider}.
Figure~\ref{tab: text-to-SQL-framework} shows an example.
Given a user utterance ``\emph{What are the major cities in the state of Kansas?''}, the system outputs a corresponding SQL that can be used for retrieving the answer from a database.
It builds a natural language interface to the database (NLIDB) to help lay users access information in the database~\cite{popescu2003towards, li2014constructing}, inspiring research in human-computer interaction~\cite{elgohary2020speak}. Because the SQL query can be regarded as a semantic representation~\cite{guo2020benchmarking}, text-to-SQL is also a representative task in semantic parsing, helping downstream applications such as question answering~\cite{wang2020text}. 
Thus, text-to-SQL has attracted researchers in the natural language processing (NLP) and the database (DB) community for decades~\cite{DBLP:journals/cacm/Codd70, hemphill1990atis, dahl1994expanding, zelle1996learning, popescu2003towards, uszkoreit2006contextual, wang2019rat, scholak2021picard}.

\begin{table*}[ht!]
\small
\centering
\begin{tabular}{lrrrrC{2.1cm}L{2.6cm}}
\toprule
Datasets & \#Size &  \#DB & \#D & \lcell{\#T/DB} & Issues addressed & Sources for data\\
\toprule

Spider~\cite{yu2018spider}      & 10,181 & 200    & 138 & 5.1  & Domain generalization & College courses, DabaseAnswers, WikiSQL\\

\rg WikiSQL~\cite{zhong2017seq2sql}    & 80,654 & 26,521 & -   & 1  & Data size & Wikipedia\\

 Squall~\cite{shi2020potential}      & 11,468 & 1,679  & -   & 1  & Lexicon-level supervision & WikiTableQuestions\\

\rg KaggleDBQA~\cite{lee2021kaggledbqa}  & 272    & 8      & 8   & 2.3 & Domain generalization  & Real web daabases\\
 
\midrule

 IMDB~\cite{yaghmazadeh2017sqlizer}        & 131    & 1      & 1   & 16  & - & Internet Movie Database \\

\rg Yelp~\cite{yaghmazadeh2017sqlizer}        & 128    & 1      & 1   & 7   & - & Yelp website \\

Advising~\cite{finegan2018improving}    & 3,898  & 1      & 1   & 10  & - & University of Michigan course information \\

\rg MIMICSQL~\cite{wang2020text}    & 10,000 & 1      & 1   & 5   & - & Healthcare domain\\

 SEDE~\cite{hazoom2021text}        & 12,023 & 1      & 1   & 29  & SQL template diversity & Stack Exchange\\
 
 \bottomrule
 \end{tabular}
 \caption{The statistic for recent text-to-SQL datasets. \#Size, \#DB, \#D, and \#T/DB represent the numbers of question-SQL pairs, databases, domains, and the averaged number of tables per domain, respectively. The ``-'' in the  \#D column indicates an unknown number of domains, and the ``-'' in the Issues Addressed indicates no specific issue addressed by the dataset. Datasets above and below the line are cross-domain and single-domain, respectively. 
The complete statistic is listed in Table~\ref{tab: dataset-stats} in Appendix~\ref{sec: text-to-SQL-datasets}.}
\label{tab: brief-dataset-stats}
\end{table*}

The challenges in text-to-SQL lie within three aspects: (1) extracting the meaning of natural utterances (encoding); (2) transforming the extracted meaning into another expression which is pragmatically equivalent to the NL meaning (translating) and; (3) producing the corresponding SQL queries (decoding).
A wide range of methods has been investigated to address the technical challenges, from representation learning, intermediate structures, decoding, model structures, training objectives, and other perspectives. 
In addition, much work has been conducted on data resources and evaluation. 
However, relatively little work has been done in the literature to provide a comprehensive survey of the landscape.
The only exceptions are~\cite{katsogiannis2021deep} and \cite{kalajdjieski2020recent}, but they cover a limited scope. To this end, we aim to provide a systematic survey that involves a broader range of text-to-SQL research and addresses the aforementioned challenges.

In this paper, we survey the recent progress on text-to-SQL, from datasets (\S~\ref{sec: datasets}), methods (\S~\ref{sec: methodologies}) to evaluation (\S~\ref{sec: evaluations})~\footnote{Note that most  work discussed in this paper is in English unless otherwise specified.} and highlight potential directions for future work (\S~\ref{sec: discussion}).
Apendix~\ref{sec: topology-for-text-to-SQL} shows the topology for the text-to-SQL task.


\section{Datasets}
\label{sec: datasets}

As shown in Table~\ref{tab: brief-dataset-stats}, existing text-to-SQL datasets can be classified into three categories: single-domain datasets, cross-domain datasets and others.

\paragraph{Single-Domain Datasets}
Single-domain text-to-SQL datasets typically collect question-SQL pairs for a single database in some real-world tasks, including early ones such as Academic~\cite{li2014constructing}, Advising~\cite{finegan2018improving}, ATIS~\cite{price1990evaluation, dahl1994expanding}, GeoQuery~\cite{zelle1996learning}, Yelp and IMDB~\cite{yaghmazadeh2017sqlizer}, Scholar~\cite{iyer2017learning} and Restaurants~\cite{tang2000automated, popescu2003towards}, as well as recent ones such as  SEDE~\cite{hazoom2021text}, ESQL~\cite{chen2021leveraging} and MIMICSQL~\cite{wang2020text}.

These single-domain datasets, particularly the early ones, are usually limited in size, containing only a few hundred to a few thousand examples.
Because of the limited size and similar SQL patterns in training and testing phases, text-to-SQL models that are trained on these single-domain datasets can achieve decent performance by simply memorizing the SQL patterns and fail to generalize to unseen SQL queries or SQL queries from other domains~\cite{finegan2018improving, yu2018spider}. 
However, since these datasets are adapted from real-life applications, most of them contain domain knowledge~\cite{gan2021exploring} and dataset conventions~\cite{suhr2020exploring}.
Thus, they are still valuable to evaluate models' ability to generalize to new domains and explore how to incorporate domain knowledge and dataset convention to model predictions.

Appendix~\ref{sec: text-to-sql-examples} gives a detailed discussion on domain knowledge and dataset convention, and concrete text-to-SQL examples.

\paragraph{Large Scale Cross-domain Datasets}
Large cross-domain datasets such as WikiSQL~\cite{zhong2017seq2sql} and Spider~\cite{yu2018spider} are proposed to better evaluate deep neural models. 
WikiSQL uses tables extracted from Wikipedia and lets annotators paraphrase questions generated for the tables. Compared to other datasets, WikiSQL is an order of magnitude larger, containing 80,654 natural utterances in total~\cite{zhong2017seq2sql}. However, WikiSQL contains only simple SQL queries, and only a single table is queried within each SQL query~\cite{yu2018spider}.

\citet{yu2018spider} propose Spider, which contains 200 databases with an average of 5 tables for each database, to test models' performance on complicated unseen SQL queries and their ability to generalize to new domains. Furthermore, researchers expand Spider to study various issues of their interest~\cite{lei2020re,zeng2020photon, gan2021exploring, taniguchi2021investigation, gan2021towards}.

Besides, researchers build several large-scale text-to-SQL datasets in different languages such as CSpider~\cite{min2019pilot}, TableQA~\cite{sun2020tableqa}, DuSQL~\cite{wang2020dusql} in Chinese, ViText2SQL~\cite{nguyen2020pilot} in Vietnamese, and PortugueseSpider~\cite{jose2021mrat} in Portuguese.
Given that human translation has shown to be more accurate than machine translation~\cite{min2019pilot}, these datasets are annotated mainly by human experts based on the English Spider dataset. 
These Spider-based datasets can serve as potential resources for multi-lingual text-to-SQL research.

\paragraph{Other Datasets}
\label{sec: other-datasets}
Several context-dependent text-to-SQL datasets have been proposed, which involve user interactions with the text-to-SQL system in English~\cite{price1990evaluation, dahl1994expanding, yu2019cosql, yu2019sparc} and Chinese~\cite{guo2021chase}.
In addition, researchers collect datasets to study questions in text-to-SQL being answerable or not~\cite{zhang2020did}, lexicon-level mapping~\cite{shi2020potential} and cross-domain evaluation for real Web databases~\cite{lee2021kaggledbqa}.

Appendix~\ref{subsec: other-text-to-SQL-datasets} discusses more details about datasets mentioned in \S~\ref{sec: other-datasets}.


\section{Methods}
\label{sec: methodologies}


Early text-to-SQL systems employ rule-based and template-based methods~\cite{li2014constructing, mahmud2015rule}, which is suitable for simple user queries and databases.
However, with the progress in both DB and NLP communities, recent work focuses on more complex settings~\cite{yu2018spider}.
In these settings, deep models can be more useful because of their great feature representation ability and generalization ability.

In this survey, we focus on the deep learning methods primarily. We divide these methods employed in text-to-SQL research into Data Augmentation~(\S~\ref{subsec: data-augmentation}), Encoding~(\S~\ref{subsec: encoding}), Decoding~(\S~\ref{subsec: decoding}), Learning Techniques~(\S~\ref{subsec: learning-techniques}), and Miscellaneous~(\S~\ref{subsec: miscellaneous}).

\subsection{Data Augmentation}
\label{subsec: data-augmentation}

Data augmentation can help text-to-SQL models handle complex or unseen questions~\cite{zhong2020grounded, wang2021learning}, achieve state-of-the-art with less supervised data~\cite{guo2018question}, and attain robustness towards different types of questions~\cite{radhakrishnan2020colloql} .

Typical data augmentation techniques involve paraphrasing questions and filling pre-defined templates for increasing data diversity.~\citet{iyer2017learning} use the Paraphrase Database (PPDB)~\cite{ganitkevitch2013ppdb} to generate paraphrases for training questions. Appendix~\ref{sec: text-to-sql-examples} gives an example of this augmentation method.~\citet{iyer2017learning} and~\citet{ yu2018syntaxsqlnet} collect question-SQL templates and fill in them with DB schema. Researchers also employ neural models to generate natural utterances for sampled SQL queries to acquire more data. For instance,~\citet{li2020seqgensql} fine-tune pre-trained T5 model~\cite{raffel2019exploring} using SQL query as the input to predict natural utterance on WikiSQL, and then randomly synthesize SQL queries from tables in WikiSQL and use the tuned model to generate the corresponding natural utterance.

The quality of the augmented data is important because low-quality data can hurt the performance of the models~\cite{zhang2021data}. Various approaches have been exploited to improve the quality of the augmented data. After sampling SQL queries,~\citet{zhong2020grounded} employ an utterance generator to generate natural utterances and a semantic parser to convert the generated natural utterance to SQL queries. 
To filter out low-quality augmented data,~\citet{zhong2020grounded} only keep data that have the same generated SQL queries as the sampled ones.~\citet{zhang2021data} use a hierarchical SQL-to-question generation process to obtain high-quality data. Observing that there is a strong segment-level mapping between SQL queries and natural utterances,~\citet{zhang2021data} decompose SQL queries into several clauses, translate each clause into a sub-question, and then combine the sub-questions into a complete question.

To increase the diversity of the augmented data,~\citet{guo2018question} incorporate a latent variable in their SQL-to-text model to encourage question diversity.~\citet{radhakrishnan2020colloql} augment the WikiSQL dataset by simplifying and compressing questions to simulate the colloquial query behavior of end-users. ~\citet{wang2021learning} exploit a probabilistic context-free grammar (PCFG) to explicitly model the composition of SQL queries, encouraging sampling compositional SQL queries.

\begin{table}[t!]
\small
\centering
\begin{tabular}{lL{2.5cm}C{2cm}}
\toprule
Methods & Adopted by & Applied datasets \\
\toprule

 Encode type&  TypeSQL~\cite{yu2018typesql} & WikiSQL \\

\rg Graph-based & GNN~\cite{bogin2019representing} & Spider \\
 
 Self-attention &  RAT-SQL~\cite{wang2019rat} & Spider \\
 
\rg Adapt PLM & SQLova~\cite{hwang2019comprehensive} & WikiSQL \\
 
 Pre-training & TaBERT~\cite{yin2020tabert} & Spider \\

 \bottomrule
\end{tabular}
\caption{Typical methods used for encoding in text-to-SQL. The full table of existing methods and more details are listed in Table~\ref{tab: encoding} in Appendix~\ref{sec: encoding-decoding-medthod}.}
\label{tab: brief-encoding}
\end{table}

\subsection{Encoding}
\label{subsec: encoding}

Various methods have been adopted to address the challenges of representing the meaning of questions, representing the structure for DB schema, and linking the DB content to question.
We group them into five categories, as shown in Table~\ref{tab: brief-encoding}.

\paragraph{Encode Token Types}

To better encode keywords such as entities and numbers in questions,~\citet{yu2018typesql} assign a type to each word in the question, with a word being an entity from the knowledge graph, a column, or a number.~\citet{yu2018spider} concatenate word embeddings and the corresponding type embeddings to feed into their model.

\paragraph{Graph-based Methods} Since DB schemas contain rich structural information, graph-based methods are used to better encode such structures.

As summarized in \S~\ref{sec: datasets}, datasets prior to Spider typically involve simple DBs that contain only one table or a single DB in both training and testing. As a result, modeling DB schema receives little attention. Because Spider contains complex and different DB in training and testing,~\citet{bogin2019representing} propose to use graphs to represent the structure of the DB schemas. Specifically,~\citet{bogin2019representing} use nodes to represent tables and columns, edges to represent relationships between tables and columns, such as tables containing columns, primary key, and foreign key constraints, and then use graph neural networks (GNNs)~\cite{li2015gated} to encode the graph structure. In their subsequent work, \citet{bogin2019global} use a graph convolutional network (GCN) to capture DB structures and a gated GCN to select the relevant DB information for SQL generation. RAT-SQL~\cite{wang2019rat} encodes more relationships for DB schemas such as ``both columns are from the same table'' in their graph.

Graphs have also been used to encode questions together with DB schema. Researchers have been using different types of graphs to capture the semantics in NL and facilitate linking between NL and table schema.~\citet{cao2021lgesql} adopt line graph~\cite{gross2018graph} to capture multi-hop semantics by meta-path (e.g., an exact match for a question token and column, together with the column belonging to a table can form a 2-hop meta-path) and distinguish between local and non-local neighbors so that different tables and columns will be attended differently. SADGA~\cite{cai2021sadga} adopts the graph structure to provide a unified encoding for both natural utterances and DB schemas to help question-schema linking. Apart from the relations between entities in both questions and DB schema, the structure for DB schemas, $\text{S}^2$SQL~\cite{hui2022s} integrates syntax dependency among question tokens into the graph to improve model performance. To improve the generalization of the graph method for unseen domains, ShawdowGNN~\cite{chen2021shadowgnn} ignores names of tables or columns in the database and uses abstract schemas in the graph projection neural network to obtain delexicalized representations of questions and DB schemas.

Finally, graph-based techniques are also exploited in context-dependent text-to-SQL. For instance, IGSQL~\cite{cai2020igsql} uses a graph encoder to utilize historical information of DB schemas in the previous turns.

\paragraph{Self-attention}

Models using transformer-based encoder~\cite{he2019x, hwang2019comprehensive, xie2022unifiedskg} incorporate the original self-attention mechanism by default because it is the building block of the transformer structure.

RAT-SQL~\cite{wang2019rat} applies relation-aware self-attention, a modified version of self-attention~\cite{vaswani2017attention}, to leverage relations of tables and columns. DuoRAT~\cite{scholak2020duorat} also adopts such a relation-aware self-attention in their encoder.

\paragraph{Adapt PLM}
Various methods have been proposed to leverage the knowledge in pre-trained language models (PLMs) and better align PLM with the text-to-SQL task.
PLMs such as BERT~\cite{devlin2018bert} are used to encode questions and DB schemas. 
The modus operandi is to input the concatenation of question words and schema words to the BERT encoder~\cite{hwang2019comprehensive, choi2021ryansql}.
Other methods adjust the embeddings by PLMs. On WikiSQL, for instance, X-SQL~\cite{he2019x} replaces segment embeddings from the pre-trained encoder by column type embeddings.~\citet{guo2019content} encode two additional feature vectors for matching between question tokens and table cells as well as column names and concatenate them with BERT embeddings of questions and DB schemas. 

HydraNet~\cite{lyu2020hybrid} uses BERT to encode the question and an individual column, aligning with the tasks BERT is pre-trained on.
After obtaining the BERT representations of all columns,~\citet{lyu2020hybrid} select top-ranked columns for SQL prediction. ~\citet{liu2021awakening} train an auxiliary concept prediction module to predict which tables and columns correspond to the question. They detect important question tokens by detecting the largest drop in the confidence score caused by erasing that token in the question. Lastly, they train the PLM with a grounding module using the question tokens and the corresponding tables as well as columns. By empirical studies,~\citet{liu2021awakening} claim that their approach can awaken the latent grounding from PLM via this erase-and-predict technique.

\paragraph{Pre-training}
There have been various works proposing different pre-training objectives and using different pre-training data to better align the transformer-based encoder with the text-to-SQL task. For instance, TaBERT~\cite{yin2020tabert} uses tabular data for pre-training with objectives of masked column prediction and cell value recovery to pre-train BERT. Grappa~\cite{yu2020grappa} synthesizes question-SQL pairs over tables and pre-trains BERT with the objectives of masked language modeling (MLM) and predicting whether a column appears in the SQL query as well as what SQL operations are triggered. GAP~\cite{shi2020learning} pre-trains BART~\cite{lewis2019bart} on synthesized text-to-SQL and tabular data with the objectives of MLM, column prediction, column recovery, and SQL generation.

 \begin{table}[t!]
\small
\centering
\begin{tabular}{lL{2.5cm}C{2cm}}
\toprule
 Methods & Adopted by & Applied datasets \\
 \toprule
 Tree & SyntaxSQLNet~\cite{yu2018syntaxsqlnet} & Spider  \\
 
 \rg Sketch &  SQLNet~\cite{xu2017sqlnet} & WikiSQL \\
 
 Bottom-up & SmBop~\cite{rubin2020smbop} & Spider \\
 
 \rg Attention & \citet{wang2019learning} & WikiSQL \\
 
 Copy & ~\citet{wang2018pointing} & WikiSQL \\
 
 \rg IR & IRNet~\cite{guo2019towards} & Spider \\
 
 \multirow{2}{*}{{Others}} & Global-GCN~\citet{bogin2019global} & Spider \\
 & \citet{kelkar2020bertrand} & Spider \\
 \bottomrule
\end{tabular}
\caption{Typical methods used for decoding in text-to-SQL. The full table and more details are listed in Table~\ref{tab: decoding} in Appendix~\ref{sec: encoding-decoding-medthod}. IR: Intermediate Representation.}
\label{tab: brief-decoding}
\end{table}

\subsection{Decoding}
\label{subsec: decoding}

Various methods have been proposed for decoding to achieve a fine-grained and easier process for SQL generation and bridge the gap between natural language and SQL queries.
As shown in Table~\ref{tab: brief-decoding}, we group these methods into five main categories and other technologies.

\paragraph{Tree-based} 
Seq2Tree~\cite{dong2016language} employs a decoder that generates logical forms in a top-down manner. The components in the sub-tree are generated conditioned on their parents apart from the input question. Note that the syntax of the logical forms is implicitly learned from data for Seq2Tree. Similarly, Seq2AST~\cite{yin2017syntactic} uses an abstract syntax tree (AST) for decoding the target programming language, where the syntax is explicitly integrated with AST. Although both Seq2Tree~\cite{dong2016language} and Seq2AST~\cite{yin2017syntactic} do not study text-to-SQL datasets, their uses of trees inspire tree-based decoding in text-to-SQL. SyntaxSQLNet~\cite{yu2018syntaxsqlnet} employs a tree-based decoding method specific to SQL syntax and recursively calls modules to predict different SQL components.

\paragraph{Sketch-based}
SQLNet~\cite{xu2017sqlnet} designs a sketch aligned with the SQL grammar, and SQLNet only needs to fill in the slots in the sketch rather than predict both the output grammar and the content. Besides, the sketch captures the dependency of the predictions. Thus, the prediction of one slot is only conditioned on the slots it depends on, which avoids issues of the same SQL query with varied equivalent serializations.~\citet{dong2018coarse} decompose the decoding into two stages, where the first decoder predicts a rough sketch, and the second decoder fills in the low-level details conditioned on the question and the sketch. Such coarse-to-fine decoding has also been adopted in other works such as IRNet~\cite{guo2019towards}. To address the complex SQL queries with nested structures, RYANSQL~\cite{choi2021ryansql} recursively yields \texttt{SELECT} statements and uses a sketch-based slot filling for each of the \texttt{SELECT} statements.

\paragraph{Bottom-up} Both the tree-based and the sketch-based decoding mechanisms can be viewed as top-down decoding mechanisms.~\citet{rubin2020smbop} use a bottom-up decoding mechanism.
Given $K$ trees of height $t$, the decoder scores trees with height $t+1$ constructed by SQL grammar from the current beam, and $K$ trees with the highest scores are kept. Then, a representation of the new $K$ trees is generated and placed in the new beam.

\paragraph{Attention Mechanism}

To integrate the encoder-side information at decoding, an attention score is computed and multiplied with hidden vectors from the encoder to get the context vector, which is then used to generate an output token~\cite{dong2016language, zhong2017seq2sql}.

Variants of the attention mechanism have been used to better propagate the information encoded from questions and DB schemas to the decoder. SQLNet~\cite{xu2017sqlnet} designs column attention, where it uses hidden states from columns multiplied by embeddings for the question to calculate attention scores for a column given the question.~\citet{guo2017bidirectional} incorporate bi-attention over question and column names for SQL component selection. ~\citet{wang2019learning} adopt a structured attention~\cite{kim2017structured} by computing the marginal probabilities to fill in the slots in their generated abstract SQL queries. DuoRAT~\cite{scholak2020duorat} adopts the relation-aware self-attention mechanism in both its encoder and decoder.
Other works that use sequence-to-sequence transformer-based models or decoder-only transformer-based models incorporate the self-attention mechanism by default~\cite{scholak2021picard, xie2022unifiedskg}.

\paragraph{Copy Mechanism} Seq2AST~\cite{yin2017syntactic} and Seq2SQL~\cite{zhong2017seq2sql} employ the pointer network~\cite{vinyals2015pointer} to compute the probability of copying words from the input.~\citet{wang2018pointing} use types (e.g., columns, SQL operators, constant from questions) to explicitly restrict locations in the query to copy from and develop a new training objective to only copy from the first occurrence in the input.
In addition, the copy mechanism is also adopted in context-dependent text-to-SQL task~\cite{wang2020pg}.

\paragraph{Intermediate Representations} Researchers use intermediate representations to bridge the gap between natural language and SQL queries. IncSQL~\cite{shi2018incsql} defines actions for different SQL components and let decoder decode actions instead of  SQL queries. IRNet~\cite{guo2019towards} introduces SemQL, an intermediate representation for SQL queries that can cover most of the challenging Spider benchmark. Specifically, SemQL removes the \texttt{JOIN ON}, \texttt{FROM} and \texttt{GROUP BY} clauses, merges \texttt{HAVING} and \texttt{WHERE} clause for SQL queries. ValueNet~\cite{brunner2021valuenet} uses SemQL 2.0, which extends SemQL to include value representation. Based on SemQL, NatSQL~\cite{
gan2021natural} removes the set operators~\footnote{The operators that combine the results of two or more \texttt{SELECT} statements, such as \texttt{INTERSECT}}. ~\citet{suhr2020exploring} implement SemQL as a mapping from SQL to a representation with an under-specified \texttt{FROM} clause, which they call SQL$^{UF}$. ~\citet{rubin2020smbop} employ a relational algebra augmented with SQL operators as the intermediate representations.

However, the intermediate representations are usually designed for a specific dataset and cannot be easily adapted to others~\cite{suhr2020exploring}. To construct a more generalized intermediate representation,~\citet{herzig2021unlocking} propose to omit tokens in the SQL query that do not align to any phrase in the utterance.

Inspired by the success of text-to-SQL task, intermediate representations are also studied for SPARQL, another executable language for database systems~\cite{saparina2021sparqling, herzig2021unlocking}.

\paragraph{Others}

PICARD~\cite{scholak2021picard} and UniSAr~\cite{dou2022unisar} set constraints to the decoder to prevent generating invalid tokens.
Several methods adopt an execution-guided decoding mechanism to exclude non-executable partial SQL queries from the output candidates~\cite{wang2018robust, hwang2019comprehensive}.
Global-GNN~\cite{bogin2019global} employs a separately trained discriminative model to rerank the top-$K$ SQL queries in the decoder's output beam, which is to reason about the complete SQL queries instead of considering each word and DB schemas in isolation. Similarly,~\citet{kelkar2020bertrand} train a separate discriminator to better search among candidate SQL queries.
\citet{xu2017sqlnet, yu2018syntaxsqlnet, guo2017bidirectional, lee2019clause} use separate submodules to predict different SQL components, easing the difficulty of generating a complete SQL query.~\citet{chen2020tale} employ a gate to select between the output sequence encoded for the question and the output sequence from the previous decoding steps at each step for SQL generation.
Inspired by machine translation,~\citet{muller2019byte} apply byte-pair encoding (BPE)~\cite{sennrich-etal-2016-neural} to compress SQL queries to shorter sequences guided by AST, reducing the difficulties in SQL generation.

\subsection{Learning Techniques}
\label{subsec: learning-techniques}

Apart from end-to-end supervised learning, different learning techniques have been proposed to help text-to-SQL research. Here we summarize these learning techniques, each addressing a specific issue for the task.

\paragraph{Fully supervised}~\citet{ni2020merging} adopt {\it active learning} to save human annotation.~\citet{yao2019model, yao2020imitation, li2020you} employ {\it interactive or imitation learning} to enhance text-to-SQL systems via interactions with end-users.~\citet{huang2018natural, wang2020meta, chen2021leveraging} adopt {\it meta-learning}~\cite{finn2017model} for domain generalization. Various {\it multi-task learning} settings have been proposed to improve text-to-SQL models via enhancing their abilities on some relevant tasks. ~\citet{chang2020zero} set an auxiliary task of mapping between column and condition values. SeaD~\cite{xuan2021sead} integrates two denoising objectives to help the model better encode information from the structural data.~\citet{hui2021improving} integrate a task of learning the correspondence between questions and DB schemas.~\citet{shi2021end} integrate a column classification task to classify which columns appear in the SQL query.~\citet{mccann2018natural} and~\citet{xie2022unifiedskg} train their models with other semantic parsing tasks, which improves models' performance on text-to-SQL task.

\paragraph{Weakly supervised} Seq2SQL~\cite{zhong2017seq2sql} use {\it reinforcement learning} to learn \texttt{WHERE} clause to allow different orders for components in \texttt{WHERE} clause.~\citet{liang2018memory} leverage memory buffer to reduce the variance of policy gradient estimates when applying reinforcement learning to text-to-SQL.~\citet{agarwal2019learning} use {\it meta-learning} and {\it Bayesian optimization}~\cite{snoek2012practical} to learn an auxiliary reward to discount spurious SQL queries in SQL generation.~\citet{min2019discrete} model the possible SQL queries as a {\it discrete latent variable} and adopt a hard-{\it EM}-style parameter updates, letting their model take advantage of the possible pre-computed solutions.

\subsection{Miscellaneous}
\label{subsec: miscellaneous}

In DB linking, BRIDGE~\cite{lin2020bridging} appends a representation for the DB cell values mentioned in the question to corresponding fields in the encoded sequence, which links the DB content to the question.~\citet{ma2020mention} employ an explicit extractor of slots mentioned in the question and then link them with DB schemas.

Model-wise,~\citet{finegan2018improving} use a template-based model which copies slots from the question. ~\citet{shaw2020compositional} use a hybrid model which firstly uses a high precision grammar-based approach (NQG) to generate SQL queries, then uses T5~\cite{raffel2019exploring} as a back-up if NQG fails.~\citet{yan2020sql} formulate submodule slot-filling as machine reading comprehension (MRC) task and apply BERT-based MRC models on it. Besides, DT-Fixup~\cite{xu2020optimizing} designs an optimization approach for a deeper Transformer on small datasets for the text-to-SQL task.

In SQL generation, IncSQL~\cite{shi2018incsql} allows parsers to explore alternative correct action sequences to generate different SQL queries.~\citet{brunner2021valuenet} search values in DB to insert values into SQL query.

For context-dependent text-to-SQL, researchers adopt techniques such as turn-level encoder and copy mechanism~\cite{suhr2018learning, zhang2019editing, wang2020pg}, constrained decoding~\cite{wang2020pg}, dynamic memory decay mechanism~\cite{hui2021dynamic}, treating questions and SQL queries as two modalities, and using bi-modal pre-trained models~\cite{zheng2022hie}.


\section{Evaluation}
\label{sec: evaluations}

\paragraph{Metrics}
Table~\ref{tab: metrics} shows widely used automatic evaluation metrics for the text-to-SQL task.
Early works evaluate SQL queries by comparing the database querying results executed from the predicted SQL query and the ground-truth (or gold) SQL query~\cite{zelle1996learning, yaghmazadeh2017sqlizer} or use {\it exact string match} to compare the predicted SQL query with the gold one query~\cite{finegan2018improving}. However, execution accuracy can create false positives for semantically different SQL queries even if they yield the same execution results~\cite{yu2018spider}. The exact string match can be too strict as two different strings can still have the same semantics~\cite{zhong2020semantic}.
Aware of these issues,~\citet{yu2018spider} adopt {\it exact set match} (ESM) in Spider, deciding the correctness of SQL queries by comparing the sub-clauses of SQL queries.~\citet{zhong2020semantic} generate databases that can distinguish the predicted SQL query and gold one. Both methods are used as official metrics on Spider.

\paragraph{Evaluation Setup}

\begin{table}[t!]
    \centering
    \small
    \begin{tabular}{L{2.6cm}L{2.6cm}C{1.1cm}}
    \toprule
    Metrics & Datasets & Errors \\
    \toprule
    Naiive Execution Accuracy & GeoQuery, IMDB, Yelp, WikiSQL, etc & False positive \\
    \rg Exact String Match & Advising, WikiSQL, etc & False negative \\
    Exact Set Match & Spider & False negative \\
    \rg Test Suite Accuracy (execution accuracy with generated databases) & Spider, GeoQuery, etc & False positive \\
    \bottomrule
    \end{tabular}
    \caption{The summary of metrics, datasets that use these metrics, and their potential error cases.}
    \label{tab: metrics}
    \vspace{-2mm}
\end{table}

Early single-domain datasets typically use the standard train/dev/test split~\cite{iyer2017learning} by splitting the question-SQL pairs randomly. To evaluate generalization to unseen SQL queries within the current domain,~\citet{finegan2018improving} propose SQL query split, where no SQL query is allowed to appear in more than one set among the train, dev, and test sets. Furthermore,~\citet{yu2018spider} propose a database split, where the model does not see the databases in the test set in its training time. Other splitting methods also exist to help different research topics~\cite{shaw2020compositional, chang2020zero}.


\section{Discussion and Future Directions}\label{sec: discussion}

Ever since the LUNAR system~\cite{lunar, woods1973progress}, systems for retrieving DB information have witnessed an increasing amount of research interest and an enormous growth, especially in the field of text-to-SQL in the deep learning era.
With the ever-increasing model performance on the WikiSQL and Spider leaderboards, one can be optimistic because models are becoming more sophisticated than ever.
But there are still several challenges to overcome.

First, these sophisticated models suffer a great performance loss when tested against different text-to-SQL datasets from other domains~\cite{suhr2020exploring, lee2021kaggledbqa}.
It is unclear how to incorporate domain knowledge to the models trained on Spider and deploy these models efficiently on different domains, especially those with similar information stored in DB but slightly different DB schemas.
Although large-scale datasets promote the cross-domain settings, question-SQL pairs from Spider are free from domain knowledge, ambiguity, or domain convention. Thus, {\it cross-domain text-to-SQL} needs to be studied in future research to build a practical cross-domain system that can handle real-world requests.

There are different {\it use cases in real-world scenarios}, which requires models to be robust to different settings and be smart to handle different user requests. For instance, the model trained with DB schemas can need to handle a corrupted table, or no table is provided in its practical use. Besides, the input from users can vary from the standard question input in Spider or WikiSQL, which poses challenges to models trained on these datasets. 
More user studies need to be done to study how well the current systems serve the end-users and the input pattern from the end-users. 
Apart from SQL queries, administrators can want to change DB schemas, where a system that can translate the natural language to such DB commands can be helpful. 
Also, although there are already works on text-to-SQL beyond English~\cite{min2019pilot, nguyen2020pilot, jose2021mrat}, but we still lack a comprehensive study on multi-lingual text-to-SQL, which can be challenging but useful in real-life scenarios.
Finally, it is important to build NLIDB for people with disabilities.~\citet{song2022speech} propose speech-to-SQL that translates voice input to SQL queries, which helps visually impaired end users. More work can be done to address various needs from the perspective of end-users, in particular, the needs from minorities.

Text-to-SQL research can also be integrated into {\it a larger scope of research}. Application-wise,~\citet{xu2020autoqa} develop a question answering system for the database,~\citet{chen2020airconcierge} generate task-oriented dialogue by retrieving knowledge from the database using the text-to-SQL model. An example of the possible directions is to employ the text-to-SQL model to query databases for fact-checking. 
Research-wise,~\citet{guo2020benchmarking} compare SQL queries to other logical forms in semantic parsing,~\citet{xie2022unifiedskg} include text-to-SQL as one of the tasks to achieve a generalized semantic parsing framework. 
The inter-relations between various logical forms in semantic parsing can be further studied. A generalized framework or a generalized model can come as the fruit for our semantic parsing community.


In hindsight, the development of text-to-SQL has been pushed by the innovation in the general ML/NLP community, such as LSTM~\cite{hochreiter1997long}, self-attention~\cite{vaswani2017attention}, PLMs~\cite{devlin2018bert}, etc. Recently, {\it prompt learning} has achieved decent performance on various tasks, in particular, in the low-resource setting~\cite{liu2021pre}. Such characteristics align well with the expectation of having a functional text-to-SQL model with a few training samples. Some recent works already explore applying prompt learning to the text-to-SQL task~\cite{xie2022unifiedskg}. The practical expectation for the text-to-SQL task is to deploy the model in different scenarios, requiring robustness across domains. However, prompt learning struggles with being robust, and the performance can be easily affected by the selected data. This misalignment encourages researchers to study how to employ prompt learning in the real-world text-to-SQL task, which can need further understanding of the cross-domain challenges for text-to-SQL.

Another line of research is to {\it evaluate these sophisticated text-to-SQL systems}. The typical measure is to evaluate the performance of the system on some existing datasets. 
As there are operational systems using NL input to perform tasks such as getting answers from database management system or building ontologies or playing some games, the performance of these systems can be measured by the diminution of the (human) time taken to get the searched information~\cite{deng2021prefix, zhou-etal-2022-online}. While there are context-dependent text-to-SQL datasets available~\cite{yu2019cosql, yu2019sparc}, researchers can draw inspirations from other fields of research~\cite{zellers-etal-2021-turingadvice} to design interactive set-ups to evaluate text-to-SQL systems. Appendix~\ref{sec: other-related-tasks} discusses tasks relevant to the task of text-to-SQL.

\section*{Acknowledgement}
Yue Zhang is the corresponding author. 
We thank all reviewers for their insightful comments, and Rada Mihalcea, Siqi Shen, Winston Wu and Ian Stewart for proofreading and suggestions.
The work is funded by the Zhejiang Province Key Project 2022SDXHDX0003. 

\bibliography{anthology,custom}
\bibliographystyle{acl_natbib}

\newpage

\appendix


\section{Topology for Text-to-SQL}
\label{sec: topology-for-text-to-SQL}
\begin{table*}[ht!]
\footnotesize
        \begin{forest}
            for tree={
                forked edges,
                grow'=0,
                draw,
                rounded corners,
                node options={align=center,},
                text width=2.7cm,
                s sep=6pt,
                calign=child edge, calign child=(n_children()+1)/2,
            },
            [text-to-SQL, fill=gray!45, parent
                [Datasets \S~\ref{sec: datasets}, for tree={ pretrain}
                    [Single-domain, pretrain
                            [ATIS; GeoQuery; Restaurants; Scholar; Academic; Yelp; IMDB; Advising; MIMICSQL; ESQL(\texttt{zh}); SEDE, 
                            style = pretrain_work]
                    ]
                    [Large Scale Cross-domain,  pretrain
                        [WikiSQL, pretrain_work]
                        [Spider; Spider-DK; Spider$ _{\text{UTran}}$; Spider-L; Spider$_\text{SL}$; Spider-Syn, pretrain_work
                        ]
                        [TableQA(\texttt{zh}); DuSQL(\texttt{zh}); ViText2SQL(\texttt{vi}); CSpider(\texttt{zh}); PortugueseSpider(\texttt{pt}), pretrain_work]
                    ]
                    [Others, pretrain
                        [Multi-turn, pretrain
                            [ATIS; Sparc; CoSQL; Splash; Chase (\texttt{zh}), pretrain_work]
                        ]
                        [Others, pretrain
                            [TriageSQL; Squall; KaggleDBQA, pretrain_work]
                        ]
                    ]
                ]
                [Methodologies \S~\ref{sec: methodologies}, for tree={fill=red!45,template}
                    [Data Augmentation, template]
                    [Encoding, template
                        [Encode Token Types; Graph-based; Self-attention; Adapt PLM; Pre-training, template_work]
                    ]
                    [Decoding, template
                        [Tree-based; Sketch-based; Bottom-up; Attention Mechanism; Copy Mechanism; Intermediate Representation; Others, template_work]
                    ]
                    [Learning Techniques, template
                        [Fully Supervised, template
                            [Active Learning; Interactive/Imitation Learning; Meta-learning; Multi-task learning, template_work]
                        ]
                        [Weakly supervised, template
                            [Reinforcement Learning; Meta-Learning; Bayesian Optimization; Hard-EM-style Parameter Updates, template_work]
                        ]
                    ]
                    [Miscellaneous, template]
                ]
                [Evaluations \S~\ref{sec: evaluations}, for tree={multiple}
                    [Metrics, multiple
                                [Exact string match; Exact set match; Execution accuracy, multiple_work]
                    ]
                    [Split Methods, multiple
                                [Example split; SQL query split; Database split, multiple_work]                          
                    ]
                ]        
            ]
        \end{forest}
            \caption{Topology for text-to-SQL. Format adapted from~\citet{liu2021pre}.}
            \label{fig:typo-text-to-SQL}
\end{table*}
Figure~\ref{fig:typo-text-to-SQL} shows the topology for the text-to-SQL task.

\section{Text-to-SQL Examples}
\label{sec: text-to-sql-examples}

\subsection{Table and Database}

Table~\ref{tab: db-table-example} shows an example of the table in the database for Restaurants dataset. The domain for this dataset is restaurant information, where questions are typically about food type, restaurant location, etc. 

There is a big difference in terms of how many tables a database has. For restaurants, there are 3 tables in the database, while there are 32 tables in ATIS~\cite{suhr2020exploring}. 

\begin{table}[ht!]
\small
\centering

\begin{tabular}{C{2.1cm}C{2.1cm}C{2.1cm}}
\toprule
CITY\_NAME* & COUNTY &  REGION \\
\midrule
VARCHAR(255) & VARCHAR(255) &  VARCHAR(255) \\
\toprule
Alameda & Alameda County & Bay Area \\
\rg Alamo & Contra Costa County & Bay Area \\
Albany & Alameda County & Bay Area \\
...&...&...\\
\bottomrule
\end{tabular}

\caption{Geography, one of the tables in Restaurants database. * denotes the primary key of this table. We only include 3 rows for demonstration purpose.}
\label{tab: db-table-example}
\end{table}

\subsection{Domain Knowledge}

\textit{Question:}
Will undergrads be okay to take 581 ?
\textit{SQL query:}
\begin{lstlisting}
SELECT DISTINCT T1.ADVISORY_REQUIREMENT , T1.ENFORCED_REQUIREMENT , T1.NAME FROM COURSE AS T1 WHERE (*@\textcolor{blue}{T1.DEPARTMENT = "EECS"}@*) AND T1.NUMBER = 581 ;
\end{lstlisting}

In Advising dataset, Department \texttt{``\textcolor{blue}{EECS}''}
 is considered as domain knowledge where ``581'' in the utterance means a course in \texttt{``EECS''} department with course number \texttt{``581''}. 

\subsection{Dataset Convention}
\label{subsec: dataset-convention}

\textit{Question:}
Give me some \textcolor{red}{restaurants} in alameda ?
\textit{SQL query:}
\begin{lstlisting}
SELECT (*@\textcolor{red}{T1.HOUSE\_NUMBER , T2.NAME}@*) FROM LOCATION AS T1 , RESTAURANT AS T2 WHERE T1.CITY_NAME = "alameda" AND T2.ID = T1.RESTAURANT_ID ;
\end{lstlisting}

In Restaurants dataset, when the user queries ``\textcolor{red}{restaurants}'', by dataset convention, the corresponding SQL query returns the column \texttt{``\textcolor{red}{HOUSE\_NUMBER}''} and \texttt{``\textcolor{red}{NAME}''}.

\subsection{Text-to-SQL Templates}
An example of the template for text-to-SQL pair used by \citet{iyer2017learning} is as follows:

\noindent\textit{Question template:}
 Get all \textcolor{blue}{<ENT1>}.\textcolor{cyan}{<NAME>} having \textcolor{blue}{<ENT2>}.\textcolor{magenta}{<COL1>}.\textcolor{cyan}{<NAME>} as \textcolor{blue}{<ENT2>}.\textcolor{magenta}{<COL1>}.\textcolor{violet}{<TYPE>}
 
\noindent\textit{SQL query template:}
\begin{lstlisting}
SELECT (*@\textcolor{blue}{<ENT1>}.\textcolor{magenta}{<DEF>}@*) FROM JOIN_FROM((*@\textcolor{blue}{<ENT1>, <ENT2>}@*)) WHERE JOIN_WHERE((*@\textcolor{blue}{<ENT1>, <ENT2>}@*)) AND
(*@\textcolor{blue}{<ENT2>}.\textcolor{magenta}{<COL1>} = \textcolor{blue}{<ENT2>}.\textcolor{magenta}{<COL1>}.\textcolor{violet}{<TYPE>}@*) ;
\end{lstlisting}

\noindent\textit{Generated question:} Get all \textcolor{blue}{author} having \textcolor{blue}{dataset} as \textcolor{violet}{DATASET\_TYPE}

\noindent\textit{Generated SQL query:}
\begin{lstlisting}
SELECT (*@\textcolor{blue}{author}.\textcolor{magenta}{authorId}@*)
FROM author , writes , paper , paperDataset , dataset WHERE author.authorId = writes.authorId
AND writes.paperId = paper.paperId
AND paper.paperId = paperDataset.paperId AND paperDataset.datasetId = dataset.datasetId AND (*@\textcolor{blue}{dataset}.\textcolor{magenta}{datasetName}@*) = (*@\textcolor{violet}{DATASET\_TYPE}@*) ;
\end{lstlisting}
, where they populate the slots in the templates with table and column names from the database schema, as well as join the corresponding tables accordingly.

An example of the PPDB~\cite{ganitkevitch2013ppdb} paraphrasing is ``thrown into jail'' and ``imprisoned''. The English portion of PPDB contains over 220 million paraphrasing pairs.

\subsection{Complexity of Natural Language and SQL Query Pairs}

In terms of the complexity for SQL queries, \citet{finegan2018improving} find that models perform better on shorter SQL queries than longer SQL queries, which indicates that shorter SQL queries are easier in general.
\citet{yu2018spider} define the SQL hardness as the number of SQL components. The SQL query is harder when it contains more SQL keywords such as \texttt{\textcolor{blue}{GROUP BY}} and nested subqueries.~\citet{yu2018spider} gives some examples of SQL queries with different difficulty levels:

\noindent\textit{Easy:}

\begin{lstlisting}
SELECT COUNT(*)
FROM cars_data
WHERE cylinders > 4 ;
\end{lstlisting}

\noindent\textit{Medium:}

\begin{lstlisting}
SELECT T2.name, COUNT(*)
FROM concert AS T1 (*@\textcolor{blue}{JOIN}@*) stadium AS T2 ON T1.stadium_id = T2.stadium_id (*@\textcolor{blue}{GROUP BY}@*) T1.stadium_id ;
\end{lstlisting}

\noindent\textit{Hard:}

\begin{lstlisting}
SELECT T1.country_name
FROM countries AS T1 (*@\textcolor{blue}{JOIN}@*) continents AS T2 ON T1.continent = T2.cont_id (*@\textcolor{blue}{JOIN}@*) car_makers AS T3 ON T1.country_id = T3.country
WHERE T2.continent = 'Europe'
(*@\textcolor{blue}{GROUP BY}@*) T1.country_name
(*@\textcolor{blue}{HAVING}@*) COUNT(*) >= 3 ;
\end{lstlisting}

\noindent\textit{Extra Hard:}

\begin{lstlisting}
SELECT AVG(life_expectancy) FROM country
WHERE name (*@\textcolor{blue}{NOT IN}@*)
   (*@\textcolor{blue}{(SELECT T1.name}@*)
    (*@\textcolor{blue}{FROM country AS T1 JOIN}@*)
    (*@\textcolor{blue}{country\_language AS T2}@*)
    (*@\textcolor{blue}{ON T1.code = T2.country\_code}@*)
    (*@\textcolor{blue}{WHERE T2.language = "English"}@*)
      (*@\textcolor{blue}{AND T2.is\_official = "T")}@*) ;
\end{lstlisting}

In terms of the complexity of natural utterance, there is no qualitative measure of how hard the utterance is. Intuitively, models' performance can decrease when faced with longer questions from users. However, the information conveyed in longer sentences can be more complete, while there can be ambiguity in shorter sentences. Besides, there can be domain-specific phrases that confuse the model in both short and long utterances~\cite{suhr2020exploring}. Thus, researchers need to consider various perspectives to determine the complexity of natural utterance.

\section{Text-to-SQL Datasets}
\label{sec: text-to-SQL-datasets}

Table~\ref{tab: dataset-stats} lists statistics for text-to-SQL datasets.

\begin{table*}[ht!]
\small
\centering
\begin{tabular}{lrrrrC{2.1cm}L{2.6cm}}
\toprule
Datasets & \#Size & \#DB & \#D & \lcell{\#T/DB} & Issues addressed & Sources for data\\
\toprule

Spider~\cite{yu2018spider}      & 10,181 & 200    & 138 & 5.1  & Domain generalization & College courses, \href{http://www.databaseanswers.org/}{DabaseAnswers}, WikiSQL\\

\rg Spider-DK~\cite{gan2021exploring}  & 535    & 10     & -   & 4.8 & Domain knowledge & Spider dev set \\

Spider$_{\text{Utran}}$~\cite{zeng2020photon} & 15,023 & 200    & 138 & 5.1 & Untranslatable questions & Spider + 5,330 untranslatable questions \\

\rg Spider-L~\cite{lei2020re} & 8,034 & 160 & - & 5.1 & Schema linking & Spider train/dev \\

 Spider$_{\text{SL}}$~\cite{taniguchi2021investigation} & 1,034 & 10 & - & 4.8 & Schema linking & Spider dev set \\

\rg Spider-Syn~\cite{gan2021towards} & 8,034 & 160 & - & 5.1& Robustness & Spider train/dev\\

WikiSQL~\cite{zhong2017seq2sql}    & 80,654 & 26,521 & -   & 1  & Data size & Wikipedia\\

\rg Squall~\cite{shi2020potential}      & 11,468 & 1,679  & -   & 1  & Lexicon-level supervision & WikiTableQuestions \cite{pasupat2015compositional}\\

 KaggleDBQA~\cite{lee2021kaggledbqa}  & 272    & 8      & 8   & 2.3 & Domain generalization  & Real web daabases\\

\midrule

\rg ATIS~\cite{price1990evaluation, dahl1994expanding}        & 5,280  & 1      & 1   & 32 & - & Flight-booking \\

 GeoQuery~\cite{zelle1996learning}    & 877    & 1      & 1   & 6  & - & US geography  \\
 
\rg Scholar~\cite{iyer2017learning}     & 817    & 1      & 1   & 7   & - & Academic publications \\

 Academic~\cite{li2014constructing}    & 196    & 1      & 1   & 15  & - & Microsoft Academic Search (MAS) database \\

\rg IMDB~\cite{yaghmazadeh2017sqlizer}        & 131    & 1      & 1   & 16  & - & Internet Movie Database \\

 Yelp~\cite{yaghmazadeh2017sqlizer}        & 128    & 1      & 1   & 7   & - & Yelp website \\

\rg Advising~\cite{finegan2018improving}    & 3,898  & 1      & 1   & 10  & - & University of Michigan course information \\

 \lcell{\gsrow{Restaurants~\cite{tang2000automated}}\\~\cite{popescu2003towards}} & 378    & 1      & 1   & 3  & - & Restaurants  \\

\rg MIMICSQL~\cite{wang2020text}    & 10,000 & 1      & 1   & 5   & - & Healthcare domain\\

 SEDE~\cite{hazoom2021text}        & 12,023 & 1      & 1   & 29  & SQL template diversity & Stack Exchange\\
 \bottomrule
 \end{tabular}
 \caption{Summarization for text-to-SQL datasets. \#Size, \#DB, \#D, and \#T/DB represent the number of question-SQL pairs, databases, domains, and tables per domain, respectively. We put ``-'' in the \#D column because we do not know how many domains are in the Spider dev set and ``-'' in the Issues Addressed column because there is no specific issue addressed for the dataset. Datasets above and below the line are cross-domain and single-domain, respectively.}
\label{tab: dataset-stats}
\end{table*}

\subsection{More Discussion on Text-to-SQL Datasets}
\label{subsec: other-text-to-SQL-datasets}

CSpider~\cite{min2019pilot}, ViText2SQL~\cite{nguyen2020pilot} and~\citet{jose2021mrat} translate all the English questions in Spider into Chinese, Vietnamese and Portuguese, respectively. TableQA~\cite{sun2020tableqa} follows the data collection method from WikiSQL, while DuSQL~\cite{wang2020dusql} follows Spider. Both TableQA and DuSQL collect Chinese utterance and SQL query pairs across different domains.~\citet{chen2021leveraging} propose a Chinese domain-specific dataset, ESQL. 

For multi-turn context-dependent text-to-SQL benchmarks, ATIS~\cite{price1990evaluation, dahl1994expanding} includes user interactions with a SQL flight database in multiple turns. Sparc~\cite{yu2019sparc} takes a further step to collect multi-turn interactions across 200 databases and 138 domains. However, both ATIS and Sparc assume all user questions can be mapped into SQL queries and do not include system responses. Later, inspired by task-oriented dialogue system~\cite{budzianowski2018multiwoz},~\citet{yu2019cosql} propose CoSQL. In CoSQL, the dialogue state is tracked by SQL. CoSQL includes three tasks of SQL-grounded dialogue state tracking to generate SQL queries from user's utterance, system response generation from query results, and user dialogue act prediction to detect and resolve ambiguous and unanswerable questions.

Besides, TriageSQL~\cite{zhang2020did} collects unanswerable questions other than natural utterance and SQL query pairs from Spider and WikiSQL, bringing up the challenge of distinguishing answerable questions from unanswerable ones in text-to-SQL systems.

\section{Encoding and Decoding Method}
\label{sec: encoding-decoding-medthod}

\begin{table*}[ht!]
\small
\centering
\begin{tabular}{lL{5cm}C{2cm}L{3cm}}
\toprule
Methods & Adopted by & Applied datasets & Addressed challenges\\
\toprule
 Encode token type&  TypeSQL~\cite{yu2018typesql} & WikiSQL & Representing question meaning \\

\rg & GNN~\cite{bogin2019representing} & Spider & \CC{}\\
\rg & Global-GCN~\cite{bogin2019global} & Spider & \CC{}\\
\rg & IGSQL~\cite{cai2020igsql}&  Sparc, CoSQL & \CC{}\\
\rg & RAT-SQL~\cite{wang2019rat} & Spider & \CC{}\\
\rg & LEGSQL~\cite{cao2021lgesql} & Spider & \CC{}\\
\rg & SADGA~\cite{cai2021sadga} & Spider & \CC{}\\
\rg & ShawdowGNN~\cite{chen2021shadowgnn} & Spider & \CC{}\\
\rg  \multirow{-8}{*}{Graph-based} & $\text{S}^2$SQL~\cite{hui2022s} & Spider, Spider-Syn & \CC{} \\

 \multirow{5}{*}{Self-attention} & X-SQL~\cite{he2019x} & WikiSQL & \CC{}\\
 & SQLova~\cite{hwang2019comprehensive} & WikiSQL & \CC{} \\
 & RAT-SQL~\cite{wang2019rat} & Spider &  \CC{}\\
 & DuoRAT~\cite{scholak2020duorat} & Spider & \CC{}\\
 & UnifiedSKG~\cite{xie2022unifiedskg} & WikiSQL, Spider &\multirow{-13}{*}{\parbox{3cm}{\CC{(1) Representing question and DB schemas in a structured way \\(2) Schema linking}}}\\
 
\rg  & X-SQL~\cite{he2019x} & WikiSQL & \WCC{}\\
\rg  & SQLova~\cite{hwang2019comprehensive} & WikiSQL & \WCC{}\\
\rg  & \citet{guo2019content} & WikiSQL & \WCC{}\\
\rg  & HydraNet~\cite{lyu2020hybrid} & WikiSQL & \WCC{}\\
\rg \multirow{-5}{*}{Adapt PLM} & \citet{liu2021awakening}, etc & Spider-L, SQUALL & \WCC{}\\
 
 & TaBERT~\cite{yin2020tabert} & Spider & \WCC{}\\
 & GraPPA~\cite{yu2020grappa} & Spider & \WCC{}\\
\multirow{-3}{*}{Pre-training} & GAP~\cite{shi2020learning} & Spider & \multirow{-8}{*}{\parbox{3cm}{\WCC{Leveraging external data to represent question and DB schemas}}}\\
 \bottomrule
\end{tabular}
\caption{Methods used for encoding in text-to-SQL.}
\label{tab: encoding}
\end{table*}

 \begin{table*}[ht!]
\small
\centering
\begin{tabular}{L{1.5cm}L{2.5cm}L{4.5cm}C{2cm}L{3cm}}
\toprule
Methods & & Adopted by & Applied datasets & Addressed challenges\\
\toprule

  && Seq2Tree~\cite{dong2016language} & - & \\
  && Seq2AST~\cite{yin2017syntactic} & - & \\
\multicolumn{2}{l}{\multirow{-3}{*}{\parbox{1.5cm}{Tree-based}}}  & SyntaxSQLNet~\cite{yu2018syntaxsqlnet} & Spider & \WCC{}\\

\rg  && SQLNet~\cite{xu2017sqlnet} & WikiSQL & \WCC{}\\
\rg  && \citet{dong2018coarse} & WikiSQL & \WCC{}\\
\rg  && IRNet~\cite{guo2019towards} & Spider & \WCC{}\\
\rg \multicolumn{2}{l}{\multirow{-4}{*}{\parbox{1.5cm}{Sketch-based}}}  & RYANSQL~\cite{choi2021ryansql} & Spider & \WCC{}\\

\multicolumn{2}{l}{Bottom-up} & SmBop~\cite{rubin2020smbop} & Spider & \multirow{-8}{*}{\WCC{\parbox{3cm}{Hierarchical decoding}}}\\

\rg && Seq2Tree~\cite{dong2016language} & - & \CC{}\\
\rg &\multirow{-2}{*}{\parbox{2.5cm}{Attention}}& Seq2SQL~\cite{zhong2017seq2sql} & WikiSQL & \CC{}\\
\cdashline{2-4}
\rg &Bi-attention& \citet{guo2017bidirectional} & WikiSQL &\CC{} \\
\cdashline{2-4}
\rg &Structured attention& \citet{wang2019learning} & WikiSQL & \CC{}\\
\cdashline{2-4}
\rg \multirow{-5}{*}{\parbox{1.5cm}{Attention Mechanism}} & Relation-aware Self-attention &
DuoRAT~\cite{scholak2020duorat} & Spider & \CC{}\\
 
  && Seq2AST~\cite{yin2017syntactic} & - & \CC{}\\
  && Seq2SQL~\cite{zhong2017seq2sql} & WikiSQL & \CC{}\\
  && ~\citet{wang2018pointing} & WikiSQL & \CC{}\\
\multicolumn{2}{l}{\multirow{-4}{*}{\parbox{1.5cm}{Copy Mechanism}}}  & SeqGenSQL~\cite{li2020seqgensql} & WikiSQL &\multirow{-9}{*}{\CC{\parbox{3cm}{Synthesizing information for decoding}}} \\

\rg && IncSQL~\cite{shi2018incsql} & WikiSQL & \WCC{}\\
\rg && IRNet~\cite{guo2019towards} & Spider & \WCC{}\\
\cdashline{3-4}
\rg && \citet{suhr2020exploring} & Spider and others$^\spadesuit$& \WCC{}\\
\cdashline{3-4}
\rg && \citet{herzig2021unlocking} & GeoQuery, ATIS, Scholar & \WCC{}\\
\cdashline{3-4}
\rg && \citet{gan2021natural} & Spider & \WCC{}\\
\rg \multicolumn{2}{l}{\multirow{-6}{*}{\parbox{1.5cm}{Intermediate Representation}}} & \citet{brunner2021valuenet} & Spider & \multirow{-6}{*}{\WCC{\parbox{3cm}{Bridging the gap between natural language and SQL query}}}\\

 &\multirow{2}{*}{\parbox{2.5cm}{Constrained decoding}}& UniSAr~\cite{dou2022unisar} & WikiSQL, Spide and others$^\heartsuit$&  \CC{}\\
  \cdashline{3-4}
 && PICARD~\cite{scholak2021picard} & Spider, CoSQL & \CC{}\\
 \cdashline{2-4}
 &\multirow{2}{*}{\parbox{2.5cm}{Execution-guided}}& SQLova~\cite{hwang2019comprehensive} & WikiSQL & \CC{}\\
 && \citet{wang2018robust} & WikiSQL &  \multirow{-4}{*}{\CC{\parbox{3cm}{Fine-grained decoding}}}\\
 \cdashline{2-4}
  &\multirow{2}{*}{\parbox{2.5cm}{Discriminative re-ranking}} & Global-GCN~\cite{bogin2019global} & Spider &  \\
 && \citet{kelkar2020bertrand} & Spider &  \multirow{-2}{*}{\parbox{3cm}{SQL Ranking}}\\
 \cdashline{2-4}
 &\multirow{3}{*}{\parbox{2.5cm}{Separate submodule}} & SQLNet~\cite{xu2017sqlnet} & WikiSQL & \CC{} \\
 && \citet{guo2017bidirectional} & WikiSQL &  \CC{}\\
 && \citet{lee2019clause} & Spider & \CC{}\\
 \cdashline{2-4}
 & BPE & \citet{muller2019byte} & Advising, ATIS, GeoQuery &  \multirow{-4}{*}{\CC{\parbox{3cm}{Easier decoding}}}\\
 \cdashline{2-4}
 \multirow{-11}{*}{\parbox{1.5cm}{Others}} & Link gating & \citet{chen2020tale} & Spider &  Synthesizing information for decoding \\
  
 \bottomrule
\end{tabular}
\caption{Methods used for decoding in text-to-SQL. $^\spadesuit$: Academic, Advising, ATIS, GeoQuery, Yelp, IMDB, Scholar, Restaurants; $^\heartsuit$: TableQA DuSQL, CoSQL, Sparc, Chase.}
\label{tab: decoding}
\end{table*}

Table~\ref{tab: encoding} and Table~\ref{tab: decoding} show the encoding and decoding methods that have been discussed in \S~\ref{subsec: encoding} and \S~\ref{subsec: decoding}, respectively.

\section{Other Related Tasks}
\label{sec: other-related-tasks}
Other tasks that are related to text-to-SQL include text-to-python~\cite{bonthu2021text2pycode}, text-to-shell script/bash script~\cite{bharadwaj2022efficient}, text-to-regex~\cite{ye2020sketch}, text-to-SPARQL~\cite{ochieng2020parot}, etc. 
They all take natural language queries as input and output different logical forms.
Among these tasks, text-to-SPARQL is closest to text-to-SQL as both SPARQL and SQL can execute on database systems. 
Therefore, some end-to-end models that take user queries as the input and output a sequence of logical forms can be applied to both tasks~\cite{raffel2019exploring}. 
In contrast, methods~\cite{xu2017sqlnet} designed to take care of SQL natures cannot be directly applied to SPARQL, which requires carefully modification instead.

\end{document}